%% file: main_icml.tex
\documentclass{article}

\usepackage{microtype}
\usepackage{graphicx}
\usepackage{subfigure}
\usepackage{booktabs} % for professional tables
\usepackage{amsmath} 

\DeclareMathOperator*{\argmin}{arg\,min}

\usepackage{hyperref}

\usepackage[accepted]{icml2018}

\icmltitlerunning{Born Again Neural Networks}

\newcommand{\vct}{\boldsymbol }
\begin{document}

\twocolumn[
\icmltitle{Born-Again Neural Networks}



\icmlsetsymbol{equal}{*}
\author{
  Tommaso Furlanello \\
  University of Southern California\\
  \texttt{furlanel@usc.edu} \\
  %% examples of more authors
  \And
  Zachary C. Lipton \\
  Carnegie Mellon University, \\
  Amazon AI \\
  %% Address \\
  \texttt{zlipton@cmu.edu} \\
  \AND
  Laurent Itti\\
  University of Southern California\\
  %% Address \\
  \texttt{itti@usc.edu} \\
  \And
  Anima Anandkumar\\
  California Institute of Technology,\\
  Amazon AI\\
  %% Address \\
   \texttt{anima@caltech.edu}\\
  %% \And
  %% Coauthor \\
  %% Affiliation \\
  %% Address \\
  %% \texttt{email} \\
}
\begin{icmlauthorlist}
\icmlauthor{Tommaso Furlanello}{usc}
\icmlauthor{Zachary C. Lipton}{cmu,ai}
\icmlauthor{Michael Tschannen}{zth}
\icmlauthor{Laurent Itti}{usc}
\icmlauthor{Anima Anandkumar}{cal,ai}
\end{icmlauthorlist}

\icmlaffiliation{usc}{University of Southern California, Los Angeles, CA, USA}
\icmlaffiliation{ai}{Amazon AI, Palo Alto, CA, USA}
\icmlaffiliation{cal}{Caltech, Pasadena, CA, USA}
\icmlaffiliation{cmu}{Carnegie Mellon University, Pittsburgh, PA, USA}
\icmlaffiliation{zth}{ETH Z\"urich, Z\"urich, Switzerland}

\icmlcorrespondingauthor{Tommaso Furlanello}{furlanel@usc.edu}

\icmlkeywords{Knowledge Distillation, Deep Learning, Metalearning, Machine Learning}

\icmlkeywords{Knowledge Distillation, Deep Learning, Metalearning, Machine Learning}

\vskip 0.3in
]

\printAffiliationsAndNotice{}  % leave blank if no need to mention equal contribution

\begin{abstract}
\input{sections/abstract.tex}
\end{abstract}

\section{Introduction}
\label{sec:intro}
\input{sections/introduction.tex}

\section{Related Literature}
\label{sec:related}
\input{sections/related.tex}
\section{Born-Again Networks}
\label{sec:ban}
\input{sections/born_again.tex}

\section{Experiments}
\label{sec:experiments}
\input{sections/experiments.tex}

\section{Results}
\label{sec:results}
\input{sections/results.tex}

\section{Discussion}
\label{sec:discussion}
\input{sections/discussion.tex}
\section*{Acknowledgements}
\label{sec:acknow}
\input{sections/acknow.tex}

\bibliography{ban_icml}
\bibliographystyle{icml2018}

% \appendix
% \section{Do \emph{not} have an appendix here}

% \textbf{\emph{Do not put content after the references.}}
% %
% Put anything that you might normally include after the references in a separate
% supplementary file.

% We recommend that you build supplementary material in a separate document.
% If you must create one PDF and cut it up, please be careful to use a tool that
% doesn't alter the margins, and that doesn't aggressively rewrite the PDF file.
% pdftk usually works fine. 

% \textbf{Please do not use Apple's preview to cut off supplementary material.} In
% previous years it has altered margins, and created headaches at the camera-ready
% stage. 
% %%%%%%%%%%%%%%%%%%%%%%%%%%%%%%%%%%%%%%%%%%%%%%%%%%%%%%%%%%%%%%%%%%%%%%%%%%%%%%%
% %%%%%%%%%%%%%%%%%%%%%%%%%%%%%%%%%%%%%%%%%%%%%%%%%%%%%%%%%%%%%%%%%%%%%%%%%%%%%%%

\end{document}

%% file: sections/abstract.tex
Knowledge Distillation (KD) consists of  transferring ``knowledge'' 
from one machine learning model (the \emph{teacher}) 
to another (the \emph{student}). 
Commonly, the teacher is a high-capacity model 
with formidable performance, 
while the student is more compact.
By transferring knowledge, 
one hopes to benefit from the student's compactness, 
without sacrificing too much performance.
We study KD from a new perspective: 
rather than compressing models, 
we train students parameterized identically to their teachers.
Surprisingly, these \emph{Born-Again Networks} (BANs),
outperform their teachers significantly, 
both on computer vision and language modeling tasks.
Our experiments with BANs based on DenseNets demonstrate state-of-the-art performance
on the CIFAR-10 (3.5\%) and CIFAR-100 (15.5\%) datasets, by validation error. 
Additional experiments explore two distillation objectives:
(i) \emph{Confidence-Weighted by Teacher Max} (CWTM) and 
(ii) \emph{Dark Knowledge with Permuted Predictions} (DKPP). Both methods
elucidate the essential components  of KD,
demonstrating the effect of the teacher outputs on both predicted and non-predicted classes.

%Anima: {let us make this more concrete. say CWTM and DKPP show improvement in accuracy and thus demonstrate the role of importance weighting in KD. we need to expand on this in the intro. It is a bit mysterious what the outcome of CWTM and DKPP experiments were, until we actually get to that section.}

%(i) \emph{dark knowledge} (DK),*  
%(ii) \emph{Confidence-Weighted by Teacher Max} (CWTM),
%and (iii) dark knowledge with permuted predictions (DKPP). 
%DK penalizes cross-entropy loss between 
%student and teacher output distributions.
%CWTM applies an importance weight equal to the teacher's argmax.
%DKPP permutes teacher's non-argmax predictions before calculating cross-entropy. 
%Our experiments 
% We present experiments 
% with students of various capacities, 
% focusing on the under-explored case where students overpower teachers.
% Our experiments show significant advantages from transferring knowledge between DenseNets and ResNets in either direction. 

%We report preliminary reports on BAN-language models improving over their teacher on Penn Tree Bank.

%% file: sections/introduction.tex
In a 2001 paper on statistical modeling \cite{breiman2001statistical},
Leo Breiman noted 
that different stochastic algorithmic procedures \citep{hansen1990neural,liaw2002classification,chen2016xgboost} 
can lead to diverse models with similar validation performances.
Moreover, he noted that we can often compose these models 
into an ensemble that achieves predictive power 
superior to each of the constituent models. 
% The same observation that multiple algorithms 
% can learn similar input-output pairings 
% can be used to decouple the learned function 
% from the specific algorithm used for deployment. 
% 
Interestingly, given such a powerful ensemble, 
one can often find a simpler model --- 
no more complex than one of the ensemble's constituents --- 
that mimics the ensemble and achieves its performance.
Previously, in \emph{Born-Again Trees} \citet{breiman1996born}
pioneered this idea, 
learning single trees that match the performance 
of multiple-tree predictors. 
These born-again trees approximate the ensemble decision 
but offer some desired properties of individual decision trees, such as their purported amenability to interpretation. 
A number of subsequent papers have proposed 
variations the idea of \emph{born-again} models.
In the neural network community,
similar ideas emerged in papers on
\emph{model compression} by \citet{bucilua2006model} 
and related work on \emph{knowledge distillation} (KD)
by \citet{hinton2015distilling}.  
In both cases, the idea is typically 
to transfer the knowledge of a high-capacity teacher 
with desired high performance 
to a more compact student \cite{ba2014deep,urban2016deep,rusu2015policy}.
Although the student cannot match the teacher 
when trained directly on the data,
the \emph{distillation} process brings the student
closer to matching the predictive power of the teacher. 

\begin{figure*}
\centering
\includegraphics[width=0.88\textwidth]{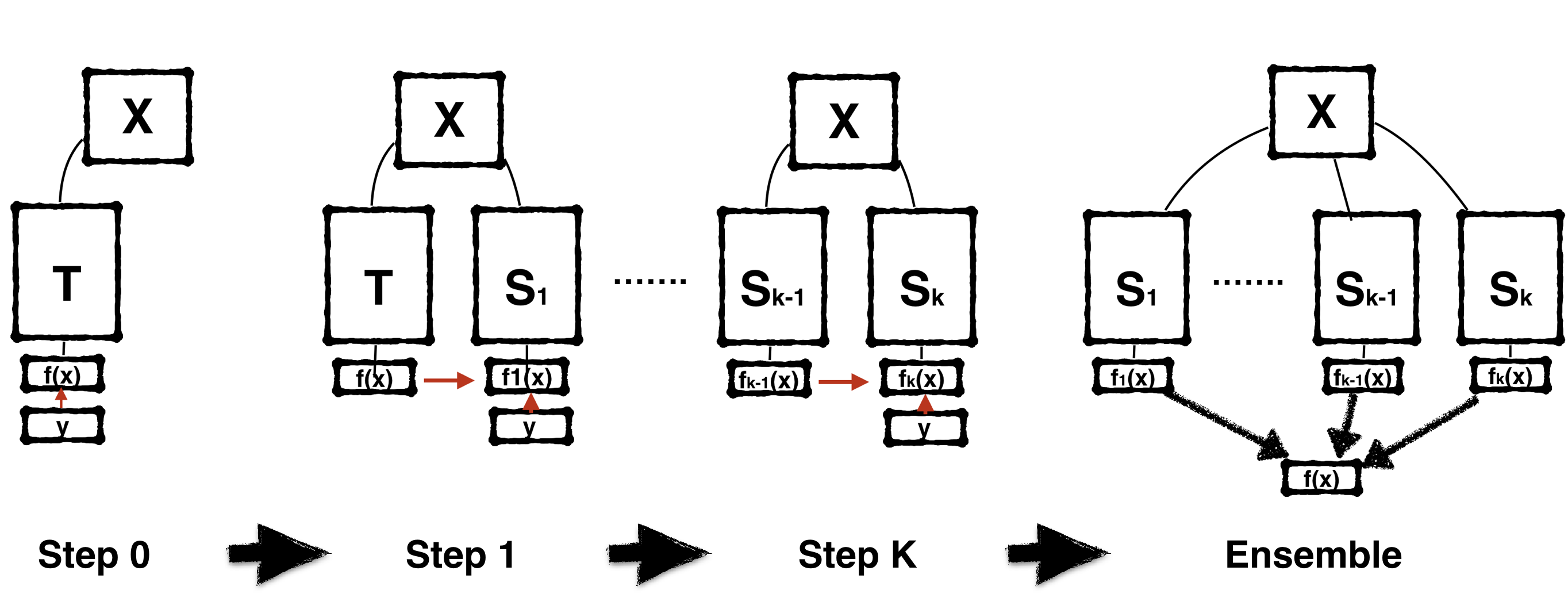}
\caption{\textbf{Graphical representation of the BAN training procedure}: during the first step the teacher model T is trained from the labels $Y$. 
Then, at each consecutive step, 
a new identical model is initialized 
from a different random seed 
and trained from the supervision of the earlier generation. At the end of the procedure,
additional gains can be achieved 
with an ensemble of multiple students generations.}
\label{fig:altm}

\end{figure*}

We propose to revisit KD
with the objective of disentangling the benefits
of this training technique from its use in model compression.
In experiments transferring knowledge 
from teachers to students of identical capacity,
we make the surprising discovery
that the students become the masters,
outperforming their teachers by significant margins.
In a manner reminiscent to Minsky's 
\emph{Sequence of Teaching Selves} \cite{minsky1991society}, 
we develop a simple re-training procedure:
after the teacher model converges, 
we initialize a new student and train it
with the dual goals of predicting the correct labels and 
matching the output distribution of the teacher.
We call these students \emph{Born-Again Networks} (BANs) 
and show that applied to DenseNets, 
ResNets and LSTM-based sequence models, 
BANs consistently have lower validation errors 
than their teachers. 
For DenseNets, we show that this procedure 
can be applied for multiple steps,
albeit with diminishing returns.

We observe that the gradient induced by KD can be decomposed into two terms: 
a \emph{dark knowledge} term, 
containing the information on the wrong outputs, 
and a ground-truth component which corresponds 
to a simple rescaling of the original gradient 
that would be obtained using the real labels. 
We interpret the second term as training 
from the real labels using importance weights 
for each sample based on the teacher's confidence 
in its maximum value. 
Experiments investigating the importance of each term are aimed at quantifying the contribution of dark knowledge
to the success of KD.

Furthermore, we explore whether the objective function induced by the DenseNet teacher can be used to improve a simpler architecture like ResNet bringing it close to state-of-the-art accuracy. 
We construct \emph{Wide-ResNets} \cite{zagoruyko2016wide} 
and \emph{Bottleneck-ResNets} \cite{he2016identity} 
of comparable complexity to their teacher
and show that these BAN-as-ResNets surpass 
their DenseNet teachers. Analogously we train DenseNet students from Wide-ResNet teachers, 
which drastically outperform standard ResNets. Thus, we demonstrate  that weak masters can still improve performance of students, and KD need not be used with    strong masters.
%here part on lstm need to be added

%% file: sections/related.tex
We briefly review the related literature
on knowledge distillation 
and the models used in our experiments. 

\subsection{Knowledge Distillation}
A long line of papers have sought
to transfer knowledge between one model 
and another for various purposes. 
Sometimes the goal is compression: 
to produce a compact model 
that retains the accuracy of a larger model 
that takes up more space and/or 
requires more computation 
to make predictions \citep{bucilua2006model,hinton2015distilling}.
\citet{breiman1996born} proposed compressing neural networks 
and multiple-tree predictors 
by approximating them with a single tree.
More recently, others have proposed 
to transfer knowledge from neural networks 
by approximating them with simpler models like decision trees \citep{chandra2007combination}
and generalized additive models \citep{tan2018transparent}
for the purpose of increasing \emph{transparency} or \emph{interpretability}.
Further, \citet{frosst2017distilling} 
proposed distilling deep networks into decision trees for the purpose of explaining decisions.
We note that in each of these cases,
what precisely is meant 
by \emph{interpretability} or \emph{transparency} is often undeclared 
and the topic remains fraught with ambiguity \citep{lipton2016mythos}.

Among papers seeking to compress models, 
the goal of knowledge transfer is simple:
produce a student model 
that achieves better accuracy 
by virtue of knowledge transfer 
from the teacher model
than it would if trained directly. 
This research is often motivated by the resource
constraints of underpowered devices 
like cellphones and internet-of-things devices.
In a pioneering work, 
\citet{bucilua2006model} compress the information 
in an ensemble of neural networks 
into a single neural network.
Subsequently, with modern deep learning tools,
\citet{ba2014deep} demonstrated a method 
to increase the accuracy of shallow neural networks, 
by training them to mimic deep neural networks,
using an penalizing the L2 norm of the difference between 
the student's and teacher's logits.
In another recent work, 
\citet{romero2014fitnets} 
aim to compress models by approximating the mappings 
between teacher and student hidden layers,
using linear projection layers 
to train the relatively narrower students.

Interest in KD increased
following \citet{hinton2015distilling}, 
who demonstrated a method called \emph{dark knowledge}, 
in which a student model trains 
with the objective of matching 
the full softmax distribution of the teacher model. 
One paper applying ML to Higgs Boson and supersymmetry detection, made the (perhaps inevitable) 
leap to applying dark knowledge to the search for dark matter \cite{sadowski2015deep}.
\citet{urban2016deep} train a \emph{super teacher}
consisting of an ensemble of 16 convolutional neural networks 
and compresses the learned function into shallow multilayer perceptrons containing 1, 2, 3, 4, and 5 layers.
In a different approach, 
\citet{zagoruyko2016paying} force the student
to match the attention map of the teacher 
(norm across the channel dimension 
in each spatial location) 
at the end of each residual stage.
\citet{czarnecki2017sobolev} try to minimize 
the difference between teacher and student derivatives 
of the loss with respect to the input 
in addition to minimizing the divergence 
from teacher predictions.

Interest in KD has also spread 
beyond supervised learning.
In the deep reinforcement learning community,
for example, 
\citet{rusu2015policy} distill multiple DQN models 
into a single one.
A number of recent papers 
\citep{furlanello2016active, 
li2016learning,shin2017continual} 
employ KD for the purpose of
minimizing forgetting in continual learning.
\cite{papernot2016distillation} incorporate KD into an adversarial training scheme.
Recently, \citet{lopez2015unifying} pointed out some connections between KD and a theory of 
on learning with privileged information \citep{pechyony2010theory}.
% \cite{pechyony2010theory} theory from Vapnik on learning with privileged information

% After developing this work, we became aware
% the most similar paper to our own:
In a superficially similar work to our own, 
\citet{yim2017gift} propose 
applying KD 
from a DNN to another DNN of identical architecture,
and report that the student model 
trains faster and achieves greater accuracy than the teacher.
They employ a loss which is calculated as follows: for a number of pairs of layers $\{(i,j)\}$ of same dimensionality, they (i) calculate a number of inner products $G_{i,j}(\vct x)$ between the activation tensors at the layers $i$ and $j$,
and (ii) they construct a loss that requires the student to match the statistics of these inner products to the corresponding statistics calculated on the teacher (for the same example), by minimizing $||G_{i,j}^T(\vct x) - G_{i,j}^S(\vct x)||^2_2$. The authors exploit a statistic used in \citet{gatys2015neural} to capture style similarity between images (given the same network).

\paragraph{Key differences}
Our work differs from \cite{yim2017gift} 
in several key ways.
First, their novel loss function,
while technically imaginative, 
is not demonstrated to outperform
more standard KD techniques.
Our work is the first, to our knowledge, to demonstrate that dark knowledge, 
applied for self-distillation,
even without softening the logits 
results in significant boosts in performance. 
Indeed, 
when distilling to a model of identical architecture
we achieve the current second-best performance on the CIFAR100 dataset. 
Moreover, this paper offers empirical rigor,
providing several experiments aimed at understanding the efficacy of self-distillation, and demonstrating that the technique is successful in domains other than images.

% First, they choose a different loss function.
% their results are considerably 
%
% This the paper at cvpr that we got pointed on twitter during nips, similar experiments drastically worse results. On you the decision of how to handle this.

% \end{itemize}
%review more results from Caruana/bengio/ attention transfer paper/iclr papers on knowledge distillation to train low precision models/policy distillation in rl/something in language/ continual learning, pseudolabels. Adversarial attacks. Label smoothing /the paper the guy told us to cite on twitter (gift from knowledge distillation).
% \begin{itemize}
% % \item \cite{breiman1996born}
% % \item \cite{bucilua2006model}
% % \item \cite{chen2015compressing}
% % \item \cite{hinton2015distilling}
% % \item \cite{sadowski2015deep}
% \end{itemize}

\subsection{Residual and Densely Connected Neural Networks}
First described in \cite{he2016deep},
deep residual networks
employ design principles 
that are rapidly becoming ubiquitous 
among modern computer vision models. 
% Resnets for image classification 
% take as input an image tensor $x$ and outputs a probability distribution over possible categories of the image. 
%First, ResNets process the input image size
%with a convolutional layer that projects 
%the 3d rgb values to a higher dimensional space (more channels).
The Resnet passes representations through 
a sequence of consecutive \emph{residual-blocks},
each of which applies several sub-modules, denoted \emph{residual units}),
each of which consists of convolutions and skip-connections,
interspersed with spatial down-sampling.
% The first convolution layer of the first unit of each block increase the channels dimensionality of the input tensor, which is then maintained constant trough the block. Each block is followed by a spatial-down-sampling operator. Finally, the output of the last block is globally-averaged and fed into a fully connected layer. 
% Following this design pattern,
% we obtain networks of different depth and complexity 
% by varying the number of units in each block 
% and/or the number of units per block. 
Multiple extensions
\cite{he2016identity,zagoruyko2016wide,xie2016aggregated,han2016deep} 
% of the original resnet hyper-parameters and blocks design 
have been proposed, 
progressively increasing their accuracy on CIFAR100 \cite{krizhevsky2009learning} and ImageNet \cite{russakovsky2015imagenet}.
Densely connected networks (DenseNets) \cite{huang2016densely} are a recently proposed variation
where the summation operation at the end of each unit 
is substituted by a concatenation of 
the input and output of the unit. 

%% file: sections/born_again.tex
Consider the classical image classification setting 
where we have a training dataset 
consisting of tuples of images and labels $(x,y)\in \mathcal{X} \times \mathcal{Y}$ 
and we are interested in fitting a function $f(x):\mathcal{X}\mapsto \mathcal{Y}$,
able to generalize to unseen data.
% an external validation set. 
Commonly, the mapping $f(x)$ is parametrized by a neural network  $f(x,\theta_{1})$, $\theta_{1}$ 
with parameters in some space $\Theta_{1}$. 
% with a carefully designed architecture encompassing known priors and desiderata over the mapping of interest.
We learn the parameters via Empirical Risk Minimization (ERM),
producing a resulting model $\theta_{1}^{*}$ 
that minimizes some loss function: 
\begin{equation}
\theta_{1}^{*}=\argmin \limits_{\theta_{1}}  \mathcal{L}(y, f(x,\theta_{1})),
\end{equation} 
typically optimized by some variant of Stochastic Gradient Descent (SGD).
% Generally, opstochastic gradient  descent (SGD) or other first order methods.

Born-Again Networks (BANs) 
are based on the empirical finding 
demonstrated in knowledge distillation / model compression papers
%that the solution $\theta_{1}^{*}$
% found by SGD can be sub-optimal 
that generalization error, 
can be reduced by modifying the loss function.
This should not be surprising:
the most common such modifications
are the classical regularization penalties
which limit the complexity of the learned model.
BANs instead exploit the idea 
demonstrated in KD,
that the information contained in 
a teacher model's output distribution $f(x,\theta_{1}^{*})$
can provide a rich source of training signal,
leading to a second solution $f(x,\theta_{2}^{*})$, $\theta_{2} \in \Theta_{2} $,  
with better generalization ability.  
We explore techniques to modify, substitute, or regularize the original loss function 
with a KD term 
based on the cross-entropy between 
the new model's outputs and the outputs
of the original model:

\begin{equation}
\mathcal{L}(f(x,\argmin \limits_{\theta_{1}}  \mathcal{L}(y, f(x,\theta_{1})) ), f(x,\theta_{2})).
\end{equation}

% Our main results are focused on the case where $\Theta_{2}$ coincides with $\Theta_{1}$,
Unlike the original works on KD, 
we address the case when 
the teacher and student networks 
have identical architectures. 
Additionally, we present experiments
addressing the case when the teacher and student networks have similar capacity but different architectures. 
For example we perform knowledge transfer from a DenseNet teacher to a ResNet student with similar number of parameters.

\subsection{Sequence of Teaching Selves Born-Again Networks Ensemble}
Inspired by  the impressive recent results of SGDR Wide-Resnet \cite{loshchilov2016sgdr}
and Coupled-DenseNet \cite{dutt2017coupled} ensembles on CIFAR100,
we apply BANs sequentially
with multiple generations of knowledge transfer.
In each case, the $k$-th model 
is trained, with knowledge transferred from the $k-1$-th student: 
\begin{equation}
\mathcal{L}(f(x,\argmin \limits_{\theta_{k-1}}  \mathcal{L}(f(x,\theta_{k-1})) ), f(x,\theta_{k})).
\end{equation}
Finally, similarly to ensembling multiple snapshots \cite{huang2017snapshot} of SGD with restart \cite{loshchilov2016sgdr}, 
we produce Born-Again Network Ensembles (BANE) by averaging the prediction of multiple generations of BANs. 
\begin{equation}
\hat{f}^{k}(x) = \sum^{k} _{i=1} f(x,\theta_{i})/k.
\end{equation}
We find the improvements of the sequence to saturate, but we are able to produce significant gains through ensembling.
% also presents re-parametrization where the student matches the size activation outputs at each size of the teacher but with different depth, and where the basic building blocks of the reborn network are switched from dense-units to residual-units.

\subsection{Dark Knowledge Under the Light}
The authors in \cite{hinton2015distilling} 
suggest that the success of KD 
depends on the dark knowledge 
hidden in the distribution of logits 
of the \emph{wrong} responses, 
that carry information on the similarity between output categories.
Another plausible explanations might be found 
by comparing the gradients flowing through output node corresponding to the correct class
during distillation 
vs. normal supervised training. 
Note that restricting attention to this gradient,
the knowledge distillation might resemble
importance-weighting where the weight corresponds to the teacher's confidence in the correct prediction.

The single-sample gradient of the cross-entropy between student logits $z_j$ and teacher logits $t_j$ with respect to the $i$th output is given by:
\begin{equation}
\dfrac{\partial \mathcal{L}_{i}}{\partial z_{i}}= q_{i}-p_{i}=\dfrac{e^{z_{i}}}{\sum\limits_{j=1}^ne^{z_{j}}}-\dfrac{e^{t_{i}}}{\sum\limits_{j=1}^{n} e^{t_{j}}}.
\end{equation}
When the target probability distribution function corresponds to  the ground truth $*$ one-hot label $p_{*}=y_{*}=1$ this reduces to:
\begin{equation}
\label{label}
\dfrac{\partial \mathcal{L}_{*}}{\partial z_{*}}= q_{*}-y_{*}=\dfrac{e^{z_{*}}}{\sum\limits_{j=1}^ne^{z_{j}}}- 1
\end{equation}

When the loss is computed 
with respect to the complete teacher output,
the student back-propagates 
the mean of the gradients 
with respect to correct and incorrect outputs 
across all the $b$ samples $s$ of the mini-batch (assuming without loss of generality the $n$th label is the ground truth label $*$):
\begin{equation}
\label{kd}
\sum\limits_{s=1}^{b}\sum\limits_{i=1}^{n}\dfrac{\partial \mathcal{L}_{i,s}}{\partial z_{i,s}}= \sum\limits_{s=1}^{b}(q_{*,s}-p_{*,s}) + \sum\limits_{s=1}^{b}\sum\limits_{i=1}^{n-1 }(q_{i,s}-p_{i,s}),
\end{equation}
up to a rescaling factor $1/b$.
The second term corresponds to the information incoming from all the wrong outputs, via dark knowledge. The first term corresponds to the gradient from the correct choice and can be rewritten as 
\begin{equation}
\label{identity}
\frac1b\sum\limits_{s=1}^{b}(q_{*,s}-p_{*,s}y_{*,s})
\end{equation}
which allows the interpretation of the output of the teacher $p_{*}$ as a weighting factor of the original ground truth label~$y_{*}$.

When the teacher is correct and confident in its output, i.e. $p_{*,s}\approx1$, Eq. \eqref{identity} reduces to the ground truth gradient in Eq. \eqref{label}, while samples with lower confidence have their gradients rescaled by  a factor $p_{*,s}$ and have reduced contribution to the overall training signal.

We notice that this form has a relationship 
with importance weighting of samples 
where the gradient of each sample in a mini-batch is balanced based on its importance weight $w_{s}$.  
When the importance weights correspond 
to the output of a teacher for the correct dimension we have:
\begin{equation}
\label{imp}
\sum\limits_{s=1}^{b}\dfrac{w_{s}}{\sum\limits_{u=1}^{b}w_{u}}(q_{*,s}-y_{*,s})=\sum\limits_{s=1}^{b}\dfrac{p_{*,s}}{\sum\limits_{u=1}^{b}p_{*,u}}(q_{*,s}-y_{*,s}).
\end{equation}

So we ask the following question: 
does the success of dark knowledge 
owe to the information contained 
in the non-argmax outputs of the teacher? 
Or is dark knowledge simply performing 
a kind of importance weighting?
To explore these questions, 
we develop two treatments.
In the first treatment, 
Confidence Weighted by Teacher Max (CWTM),
% we completely discard information on the DK term. 
we weight each example in the student's loss function (standard cross-entropy with ground truth labels)
by the confidence of the teacher model on that example (even if the teacher wrong).
We train BAN models using an approximation of Eq. \eqref{imp},
where we substitute the correct answer 
$p_{*,s}$ with the max output of the teacher 
$\max p_{.,s}$: 
\begin{equation}
\label{imp_max}
\sum\limits_{s=1}^{b}\dfrac{\max p_{.,s}}{\sum\limits_{u=1}^{b}\max p_{.,u}}(q_{*,s}-y_{*,s}).
\end{equation}
In the second treatment, 
dark knowledge with Permuted Predictions (DKPP), 
\textbf{we permute the non-argmax outputs 
of the teacher's predicted distribution}.
We use the original formulation of Eq. \eqref{kd}, 
substituting the $*$ operator with $\max$ 
and permuting the teacher dimensions 
of the dark knowledge term, leading to:
\begin{align}
\label{pkdd}
\sum\limits_{s=1}^{b}\sum\limits_{i=1}^{n}\dfrac{\partial \mathcal{L}_{i,s}}{\partial z_{i,s}}&= \sum\limits_{s=1}^{b}(q_{*,s}- \max p_{.,s}) \nonumber \\
&\qquad\quad + \sum\limits_{s=1}^{b}\sum\limits_{i=1}^{n-1 }q_{i,s}-\phi (p_{j,s}),
\end{align}
where $\phi (p_{j,s})$ are the permuted outputs of the teacher. 
In DKPP we scramble the correct attribution of dark knowledge to each non-argmax output dimension,
destroying the pairwise similarities of the original output covariance matrix.
%discuss weighted and permutation training 

\subsection{BANs Stability to Depth and Width Variations}
DenseNet architectures are parametrized by depth, growth, and compression factors. 
Depth corresponds to the number of dense blocks. 
The growth factor defines how many new features 
are concatenated at each new dense block, 
while the compression factor controls 
by how much features are reduced 
at the end of each stage.

Variations in these hyper-parameters 
induce a tradeoff between number of parameters, 
memory use and the number of sequential operations for each pass. 
We test the possibility of expressing 
the same function of the DenseNet teacher 
with different architectural hyperparameters. 
In order to construct a fair comparison,
we construct DenseNets whose output dimensionality 
at each spatial transition 
matches that of the DenseNet-90-60 teacher. 
Keeping the size of the hidden states constant,
we modulate the growth factor indirectly 
via the choice the number of blocks. 
Additionally, we can drastically reduce the growth factor by reducing the compression factor 
before or after each spatial transition.

\subsection{DenseNets Born-Again as ResNets}
Since BAN-DenseNets perform at the same level 
as plain DenseNets 
with multiples of their parameters, 
we test whether the BAN procedure 
can be used to improve ResNets as well. 
Instead of the weaker ResNet teacher, 
we employ a DenseNet-90-60 as teacher 
and construct comparable ResNet students 
by switching \emph{Dense Blocks} 
with \emph{Wide Residual Blocks} 
and \emph{Bottleneck Residual Blocks}.

%% file: sections/experiments.tex
All experiments performed on CIFAR-100 use the same preprocessing and training setting as for Wide-ResNet \cite{zagoruyko2016wide} except for  Mean-Std normalization. The only form of regularization used other than the KD loss are weight decay and, in the case of Wide-ResNet drop-out.
\subsection{CIFAR-10/100}
\paragraph{Baselines}
To get a strong teacher baseline without the prohibitive memory usage of the original architectures, 
we explore multiple heights and growth factors for DenseNets.
We find a good configuration in relatively shallower architectures with increased growth factor and comparable number of parameters to the largest configuration of the original paper. Classical ResNet baselines are trained following \cite{zagoruyko2016wide}.
Finally, we construct Wide-ResNet and bottleneck-ResNet networks 
that match the output shape 
of DenseNet-90-60 at each block, as baselines for our BAN-ResNet with DenseNet teacher experiment.

\paragraph{BAN-DenseNet and ResNet}
We perform BAN re-training after convergence, using the same training schedule
originally used to train the teacher networks.
We employ DenseNet-(116-33, 90-60, 80-80, 80-120) and train a sequence of BANs for each configuration. We test the ensemble performance for sequences of 2 and 3 BANs.
%Potentially, longer training times might benefit BAN since we found that often, the best result occurred on the final epoch of training,unlike the vanilla teacher models, which saturate earlier.
We explored other forms of knowledge transfer for training BANs. Specifically, we tried progressively constraining the BANs to be more similar to their teachers, sharing the first and last layers between student and teacher, or adding losses that penalize the L2 distance between student and teacher activations. However, we found these variations to systematically perform slightly worse than the simple KD via cross entropy. 
%Therefore, the only difference in the training setting of baselines and BAN models is the additional presence of %the knowledge distillation loss.
For BAN-ResNet experiments with a ResNet teacher we use Wide-ResNet-(28-1, 28-2, 28-5, 28-10).

\paragraph{BAN without Dark Knowledge}
In the first treatment, CWTM, we fully exclude the effect of all the teacher's output except for the argmax dimension. To do so, we train the students with the normal label loss where samples are weighted by their importance. We interpret the max of the teacher's output for each sample as the importance weight and use it to rescale each sample of the student's loss. 

In the second treatment, DKPP, we maintain the overall high order moments of the teachers output, but randomly permute each output dimension except the argmax one. We maintain the rest of the training scheme and the architecture unchanged. 

Both methods alter the covariance between outputs, such that any improvement cannot be fully attributed to the classical dark knowledge interpretation. 

\paragraph{Variations in Depth, Width and Compression Rate}
We also train variations of DenseNet-90-60, with increased or decreased number of units in each block and different number of channels determined through a ratio of the original activation sizes.

\paragraph{BAN-Resnet with DenseNet teacher}
In all the BAN-ResNet with DenseNet teacher experiments,
the student shares the first and last layers of the teacher. 
We modulate the complexity of the ResNet by changing the number of units, 
starting from the depth of the successful Wide-ResNet-28 \cite{zagoruyko2016wide} 
and reducing until only a single residual unit per block remains.
Since the number of channels in each block is the same for every residual unit, we match it with a proportion of the corresponding dense block output after the $1\times1$ convolution, before the spatial down-sampling. We explore mostly architectures with a ratio of 1, but we also show the effect of halving the width of the network.

\paragraph{BAN-DenseNet with ResNet teacher}
With this experiment we test whether a weaker ResNet teacher is able to successfully train DenseNet-90-60 students. We use multiple configurations of Wide-ResNet teacher and train the Ban-DenseNet student with the same hyper parameters of the other DenseNet experiments.

\subsection{Penn Tree Bank}
To validate our method beyond computer vision applications, we also apply the BAN framework to language models and evaluate it on the Penn Tree Bank (PTB) dataset \cite{marcus1993building} using the standard train/test/validation split by \cite{mikolov2010recurrent}.  
We consider two BAN language models: a single layer LSTM \cite{hochreiter1997long} with 1500 units \cite{zaremba2014recurrent} and a smaller model from \cite{kim2016character} combining a convolutional layers, highway layers, and a 2-layer LSTM (referred to as CNN-LSTM).

For the LSTM model we use weight tying \cite{press2016using}, 65\% dropout and train for 40 epochs using SGD with a mini-batch size of 32. An adaptive learning rate schedule is used with an initial learning rate 1 that is multiplied by a factor of 0.25 if the validation perplexity does not decrease after an epoch.

The CNN-LSTM is trained with SGD for the same number of epochs with a mini-batch size of 20. The initial learning rate is set to 2 and is multiplied by a factor of 0.5 if the validation perplexity does not decrease by at least 0.5 after an epoch (this schedule slightly differs from \cite{kim2016character}, but worked better for the teacher model in our experiments).

Both models are unrolled for 35 steps and the KD loss is simply applied between the softmax outputs of the unrolled teacher an student.

%% file: sections/results.tex
We report the surprising finding that by performing KD across models of similar architecture,
BAN student models tend to improve over their teachers across all configurations. 
\subsection{CIFAR-10}
As can be observed in Table \ref{cifar10} the CIFAR-10 test error is systematically lower or equal for both Wide-ResNet and DenseNet student trained from an identical teacher. It is worth to note how for BAN-DenseNet the gap between architectures of different complexity is quickly reduced leading to implicit gains in the parameters to error rate ratio.

\begin{table}[h]
%[!b]
\centering
\caption{\textbf{Test error on CIFAR-10} for Wide-ResNet with different depth and width and DenseNet of different depth and growth factor.}
\vspace{5pt}
\begin{tabular}{l|c|c|c}
Network           & Parameters  & Teacher & BAN\\ \hline
Wide-ResNet-28-1  & 0.38 M&  6.69 &  \textbf{6.64} \\
Wide-ResNet-28-2   & 1.48 M  &  5.06 & \textbf{4.86}  \\
Wide-ResNet-28-5   & 9.16 M & 4.13 & \textbf{4.03}   \\
Wide-ResNet-28-10  & 36 M & \textbf{3.77} & 3.86 \\
\hline
DenseNet-112-33  & 6.3 M  & 3.84 & \textbf{3.61}   \\
DenseNet-90-60   & 16.1 M  & 3.81 & \textbf{3.5}  \\
DenseNet-80-80   & 22.4 M & \textbf{3.48} & 3.49  \\
DenseNet-80-120 & 50.4 M & \textbf{3.37} & 3.54 \\

\end{tabular}
\label{cifar10}
\end{table}
\subsection{CIFAR-100}
For CIFAR-100 we find stronger improvements for all BAN-DenseNet models. We focus therefore most of our experiments to explore and understand the born-again phenomena on this dataset.

\textbf{BAN-DenseNet and BAN-ResNet}
In Table \ref{cifar100} we report test error rates using both labels and teacher outputs (BAN+L) or only the latter (BAN). The improvement of fully removing the label supervision is systematic across modality, it is worth noting that the smallest student BAN-DenseNet-112-33 reaches an error of 16.95\% with only 6.5 M parameters, comparable to the 16.87\% error of the DenseNet-80-120 teacher with almost eight times more parameters.

In Table \ref{cifar100res} all but one Wide-ResNnet student improve over their identical teacher. 
\begin{table*}[t]
%[!b]
\vspace{-0.2cm}
\centering
\caption{\textbf{Test error on CIFAR-100} \textit{Left Side: } DenseNet of different depth and growth factor and respective BAN student. BAN models are trained only with the teacher loss, BAN+L with both label and teacher loss. CWTM are trained with sample importance weighted label, the importance of the sample is determined by the max of the teacher's output. DKPP are trained only from teacher outputs with all the dimensions but the argmax permuted.
\textit{Right Side:} test error on CIFAR-100 sequence of BAN-DenseNet, and the BAN-ensembles resulting from the sequence.  Each BAN in the sequence is trained from  cross-entropy with respect to the model at its left. BAN and BAN-1 models are trained from Teacher but have different random seeds. We include the teacher as a member of the ensemble for  Ens*3 for 80-120 since we did not train a BAN-3 for this configuration.}
\vspace{5pt}
\begin{tabular}{l|c|c|c|c|c||c|c|c|c|c}
Network & Teacher & BAN & BAN+L & CWTM & DKPP &  BAN-1 & BAN-2 & BAN-3 & Ens*2 & Ens*3 \\ \hline
DenseNet-112-33    &  18.25 & \textbf{16.95}  & 17.68 &  17.84 &  17.84 & 17.61 & 17.22 & \textbf{16.59}    & 15.77 & 15.68 \\
DenseNet-90-60     & 17.69 & \textbf{16.69} & 16.93 &  17.42 & 17.43 & 16.62 & \textbf{16.44} & 16.72 & 15.39 & 15.74  \\
DenseNet-80-80   & 17.16 & \textbf{16.36} &  16.5 & 17.16 & 16.84  & 16.26 & 16.30 & \textbf{15.5}  & 15.46 & 15.14\\
DenseNet-80-120  & 16.87  & \textbf{16.00} & 16.41 & 17.12 & 16.34  & \textbf{16.13} & 16.13 & /     & \textbf{15.13} & \textbf{14.9} \\ 
\end{tabular}

\label{cifar100}
\end{table*}
\begin{table}[h]
%[!b]
\centering
\caption{\textbf{Test error on CIFAR-100} for Wide-ResNet students trained from identical Wide-ResNet teachers and for DenseNet-90-60 students trained from Wide-ResNet teachers}
\vspace{5pt}
\begin{tabular}{l|c|c|c}
Network           & Teacher & BAN & Dense-90-60\\ \hline
Wide-ResNet-28-1  & 30.05 &  29.43 & 24.93 \\
Wide-ResNet-28-2   &25.32  &  24.38 & 18.49 \\
Wide-ResNet-28-5   & 20.88 & 20.93 & 17.52   \\
Wide-ResNet-28-10  & 19.08 & 18.25 & 16.79 \\

\end{tabular}
\label{cifar100res}
\end{table}

\paragraph{Sequence of Teaching Selves}
Training BANs for multiple generations leads to inconsistent but positive improvements, that saturate after a few generations.
The third generation of BAN-3-DenseNet-80-80 produces our single best model with 22M parameters, 
achieving 15.5\% error on CIFAR0100 (Table \ref{cifar100}). 
To our knowledge, 
this is currently the SOTA non-ensemble model 
% trained with SGD
without shake-shake regularization. 
It is only beaten by \citet{anonymous2018shakedrop} who use a \emph{pyramidal ResNet} trained for 1800 epochs with a combination of shake-shake \cite{gastaldi2017shake}, pyramid-drop \cite{yamada2016deep} and cut-out regularization \cite{devries2017improved}.

\paragraph{BAN-Ensemble}
Similarly, our largest ensemble BAN-3-DenseNet-BC-80-120 with 150M parameters and an error of 14.9\% is the lowest reported ensemble result in the same setting.
BAN-3-DenseNet-112-33 is based on the building block of the best coupled-ensemble of \cite{dutt2017coupled} and reaches a single-error model of 16.59\% with only 6.3M parameters, furthermore the ensembles of two or three consecutive generations reach a comparable error of 15.77\% and 15.68\% with the baseline error of 15.68\% reported in \cite{dutt2017coupled} where four models were used.

\paragraph{Effect of non-argmax Logits}
As can be observed in the two rightmost columns if the left side of Table \ref{cifar100} we find that removing part of the dark knowledge still generally brings improvements to the training procedure with respect to the baseline. 
Importance weights CWTM lead to weak improvements over the teacher in all models but the largest DenseNet. 
Instead, in  DKPP we find a comparable but systematic improvement effect of permuting all but the argmax dimensions. 

These results demonstrate that KD does not simply contribute information on each specific non-correct output. DKPP demonstrates that the higher order moments of the output distribution that are invariant to the permutation procedure still systematically contribute to improved generalization. Furthermore, the complete removal of wrong logit information in the CWTM treatment still brings improvements for three models out of four, suggesting that the information contained in pre-trained models can be used to rebalance the training set, by giving less weight to training samples for which the teacher's output distribution is not concentrated on the max.

\paragraph{DenseNet to modified DenseNet students}
It can be seen in Table \ref{match_dense} that DenseNet students are particularly robust to the variations in the number of layers. The most shallow model with only half the number of its teacher layers DenseNet-7-1-2 still improves over the DenseNet-90-60 teacher with an error rate of 16.95\%. Deeper variations are competitive or even better than the original student. The best modified student result is 16.43\% error with twice the number of layers (half the growth factor) of its DenseNet-90-60 teacher.

The biggest instabilities as well as parameter saving is obtained by modifying the compression rate of the network, indirectly reducing the dimensionality of each hidden layer. Halving the number of filters after each spatial dimension reduction in DenseNet-14-0.5-1 gives an error of 19.83\%, the worst across all trained DenseNets. Smaller reductions lead to larger parameter savings with lower accuracy losses, but directly choosing a smaller network retrained with BAN procedure like DenseNet-106-33 seems to lead to higher parameter efficiency.
\begin{table*}[t]
\vspace{-0.2cm}
%[!b]
\centering
\caption{\textbf{Test error on CIFAR-100-Modified Densenet:} a Densenet-90-60 is used as teacher with students that share the same size of hidden states after each spatial transition but differs in depth and compression rate}
\vspace{5pt}
\begin{tabular}{l|c|c|c|c||c|c|c|l}
Densenet-90-60 & Teacher & 0.5*Depth & 2*Depth &3*Depth & 4*Depth  &  0.5*Compr & 0.75*Compr & 1.5*compr\\ \hline
Error  & 17.69 & 16.95 & 16.43 & 16.64 & 16.64 & 19.83 & 17.3 & 18.89\\
Parameters & 22.4 M & 21.2 M  &  13.7 M & 12.9 M &1 2.6 M  & 5.1 M & 10.1 M & 80.5 M \\
\end{tabular}

\label{match_dense}
\end{table*}

\begin{table*}[t]
\vspace{-0.1cm}
\centering
\caption{\textbf{DenseNet to ResNet:} CIFAR-100 test error for BAN-ResNets trained from a DenseNet-90-60 teacher with different numbers of blocks and compression factors.
In all the BAN architectures, the number of units per block is indicated first,
followed by the ratio of input and output channels with respect to a DenseNet-90-60 block. All BAN architectures share the first (conv1) and last(fc-output) layer with the teacher which are frozen. Every dense block is effectively substituted by residual blocks}
\label{match}
\vspace{5pt}
\begin{tabular}{l|c|cc}
DenseNet 90-60                                  & Parameters & Baseline                   & BAN   \\ \hline
Pre-activation ResNet-1001  & 10.2 M     & \multicolumn{1}{c|}{22.71} &   /    \\ \hline
BAN-Pre-ResNet-14-0.5                            & 7.3 M      & \multicolumn{1}{c|}{20.28} & 18.8  \\
BAN-Pre-ResNet-14-1                              & 17.7 M     & \multicolumn{1}{c|}{18.84} & \textbf{17.39 }\\ \hline
BAN-Wide-ResNet-1-1                              & 20.9 M     & \multicolumn{1}{c|}{20.4}  & 19.12 \\
BAN-Match-Wide-ResNet-2-1                              & 43.1 M     & \multicolumn{1}{c|}{18.83} & 17.42 \\
BAN-Wide-ResNet-4-0.5                            & 24.3 M     & \multicolumn{1}{c|}{19.63} & \textbf{17.13} \\
BAN-Wide-ResNet-4-1                              & 87.3 M     & \multicolumn{1}{c|}{18.77} & 17.18 \\
\end{tabular}
\label{resnet}
\end{table*}

\begin{table*}[!htbp]
\vspace{-0.1cm}
%[!b]
\centering
\caption{\textbf{Validation/Test perplexity on PTB} (lower is better) for BAN-LSTM language model of different complexity}
\vspace{5pt}
\begin{tabular}{l|c|c|c|c|c}
Network           & Parameters  & Teacher Val & BAN+L Val & Teacher Test & BAN+L Test \\ \hline
ConvLSTM  & 19M  &  83.69 &   80.27  & 80.05 & \textbf{76.97} \\
LSTM   & 52M & 75.11 & 71.19 & 71.87 & \textbf{68.56}  \\
\end{tabular}
\label{ptb}
\end{table*}

\paragraph{DenseNet Teacher to ResNet Student}
Surprisingly, we find (Ttable \ref{resnet}) 
that our Wide-ResNet and Pre-ResNet students that match the output shapes at each stage of their DenseNet teachers 
tend to outperform classical ResNets, their teachers, and their baseline. 

Both BAN-Pre-ResNet with 14 blocks per stage 
and BAN-Wide-ResNet with 4 blocks per stage 
and 50\% compression factor 
reach respectively a test error of 17.39\% and 17.13\% 
using a parameter budget that is comparable with their teachers. 
We find that for BAN-Wide-ResNets,
only limiting the number of blocks to 1 per stage leads to inferior performance compared to the teacher.

Similar to how adapting the depth of the models offers a nice tradeoff between memory consumption and number of sequential operations, exchanging dense and residual blocks allows to choose between concatenation and additions. By using additions,
ResNets overwrite old memory banks, saving RAM, at the cost of heavier models that do not share layers offering another technical tradeoff to choose from.

\paragraph{ResNet Teacher to DenseNet Students} 
The converse experiment, training a DenseNet-90-60 student from ResNet student 
confirms the trend of students 
surpassing their teachers. 
The improvement from
ResNet to DenseNet (Table \ref{cifar100res}, right-most column) 
over simple label supervision is significant
as indicated by 16.79\% error 
of the DenseNet-90-60 student 
trained from the Wide-ResNet-28-10.

\subsection{Penn Tree Bank}
Although we did not use the state-of-the-art 
bag of tricks \cite{merity2017regularizing}
for training LSTMs,
nor the recently proposed improvements on KD 
for sequence models \cite{kim2016sequence}, 
we found significant decreases in perplexity 
on both validation and testing set 
for our benchmark language models.  
The smaller BAN-LSTM-CNN model decreases test perplexity from 80.05 to 76.97, 
while the bigger BAN-LSTM model improves 
from 71.87 to 68.56. 
Unlike the CNNs trained for CIFAR classification,
we find that LSTM models work 
only when trained with a combination of teacher outputs and label loss (BAN+L).
One potential explanation for this finding
might be that teachers generally reach 100\% accuracy on the CIFAR training sets 
while the PTB training perplexity 
is far from being minimized.

%% file: sections/discussion.tex
In Marvin Minsky's Society of Mind \cite{minsky1991society},  the analysis of human development led to the idea of a \emph{sequence of teaching selves}. 
Minsky suggested that sudden spurts in intelligence during childhood may be due to longer and hidden training of new "student" models under the guidance of the older self. 
Minsky concluded
that our perception of a long-term self is constructed by an ensemble of multiple generations of internal models,
which we can use for guidance 
when the most current model falls short.
Our results show several instances where such transfer was successful in artificial neural networks.

%% file: sections/acknow.tex
This work was supported by the National Science Foundation (grant numbers CCF-1317433 and CNS-1545089), C-BRIC (one of six centers in JUMP, a Semiconductor Research Corporation (SRC) program sponsored by DARPA), and the Intel Corporation. The authors affirm that the views expressed herein are solely their own, and do not represent the views of the United States government or any agency thereof.